\documentclass[runningheads]{llncs}
\usepackage[T1]{fontenc}
\usepackage{graphicx,verbatim}
\usepackage{amsmath}
\usepackage{amssymb}
\usepackage{multirow}
\usepackage{marvosym}
\newcommand{\corr}{\mbox{\scriptsize\Letter}}

\begin{document}
\title{SIRA: Reasoning-Aware Surgical Instrument Segmentation via Query-Anchored Alignment}
\titlerunning{SIRA for Reasoning-Aware Surgical Instrument Segmentation}
%

\author{
Zhibo Zhang\inst{1} \and
Qijie Wang\inst{1} \and
Zengqiang Yan\inst{1(\corr)}
}

\authorrunning{Z. Zhang et al.}

\institute{
\textsuperscript{1}School of Electronic Information and Communications,\\
Huazhong University of Science and Technology\\
\email{\{zzb226,wangqijie,z\_yan\}@hust.edu.cn}
}
  
\maketitle              
\begin{abstract} 

Surgical instrument segmentation (SIS) plays a critical role in robotic assistance and surgical workflow analysis. However, most existing SIS methods formulate segmentation as a category-driven localization problem, limiting their ability to capture procedural context and task-dependent semantics in surgical workflows. We introduce \textbf{Reasoning-Aware Surgical Instrument Segmentation (RA-SIS)}, a task formulation that frames segmentation as query-conditioned inference under surgical context. To benchmark this setting, we construct \textbf{SurgRS}, 
a surgical reasoning segmentation dataset consisting of 41,000 image–text pairs, 
which aligns instance-level masks with structured query–answer supervision 
to enable semantic grounding at the pixel level. Based on SurgRS, we propose \textbf{Surgical Instrument Reasoning and Segmentation Assistant (SIRA)}, a multimodal framework that disentangles target-level and query-level semantics and integrates them with visual features through query-anchored dual alignment. By aligning query semantics with spatial features and segmentation prompts, SIRA enhances semantic-visual consistency in mask prediction. Extensive experiments on SurgRS demonstrate improvements over existing reasoning-aware baselines. Code is available at \url{https://github.com/linxir226/SIRA}.

\keywords{Surgical instrument segmentation \and Reasoning-aware segmentation \and Multimodal alignment.}

\end{abstract}
\section{Introduction}

Surgical Instrument Segmentation (SIS) aims to localize surgical instruments in intraoperative scenes, enabling applications such as navigation, robotic assistance, workflow analysis, and skill assessment~\cite{yue2024surgicalsam}. Most existing SIS methods formulate segmentation as a category-driven localization problem, relying primarily on visual features. Early approaches adopt CNNs~\cite{ronneberger2015u,he2017mask}, transformers~\cite{li2023transforming}, or fine-tuned large segmentation models such as Mask2Former~\cite{cheng2022masked} and SAM~\cite{yue2024surgicalsam,kirillov2023segment,ayobi2023matis}. Text-prompted extensions, including SP-SAM~\cite{yue2023surgicalpart} and TP-SIS~\cite{zhou2023text}, introduce language guidance but still treat textual input as auxiliary conditioning, with segmentation largely dominated by appearance-based matching. However, they do not explicitly ground segmentation to procedural roles or task-dependent contextual cues in surgical workflows.

Recently, reasoning-aware segmentation has emerged as a paradigm for handling context-dependent and semantically complex queries~\cite{liu2025seg}. Instead of predicting masks solely from predefined labels, models interpret structured textual instructions to infer pixel-level targets. Representative methods include LISA~\cite{lai2024lisa}, GSVA~\cite{xia2024gsva}, PathMR~\cite{zhang2025pathmr}, and GLaMM~\cite{rasheed2024glamm}. VRS-HQ~\cite{gong2025devil} pioneered the use of SAM2~\cite{ravi2024sam2} as the segmentation model, establishing itself as a robust benchmark. However, these approaches are primarily developed for natural scenes or relatively simple medical scenarios.

In real surgical workflows, instrument localization is inherently task-driven. Surgeons identify instruments by their functional role in a procedural stage, such as enabling suturing or maintaining exposure, rather than by category names~\cite{zeng2025surgvlm}. Accurate segmentation therefore requires interpreting procedural intent and contextual relationships before pixel-level prediction~\cite{long2025surgical}. This motivates a task-driven segmentation formulation.

Motivated by these observations, we introduce \textbf{Reasoning-Aware Surgical Instrument Segmentation (RA-SIS)}, which, to the best of our knowledge, is the first formulation that explicitly models surgical instrument segmentation under task-driven and context-dependent queries instead of explicit category prompts. In RA-SIS, segmentation is framed as a query-conditioned inference process, where models must first infer procedural intent and contextual semantics before producing pixel-level masks.

To benchmark RA-SIS, we construct \textbf{SurgRS}, a surgical reasoning segmentation dataset consisting of 41,000 image–text pairs that explicitly aligns refined instance-level masks with structured query–answer supervision, enabling query-conditioned pixel-level inference.

Based on SurgRS, we introduce \textbf{Surgical Instrument Reasoning and Segmentation Assistant (SIRA)}, a multimodal framework that disentangles target-level and query-level semantics and integrates them with visual features through query-anchored dual alignment. Extensive experiments on SurgRS demonstrate state-of-the-art performance.

\section{Method}

\subsection{Dataset Construction}

\subsubsection{Surgical Instrument Segmentation Dataset.}

Early benchmarks such as EndoVis-17~\cite{allan20192017} and EndoVis-18~\cite{allan20202018} provide pixel-level masks for surgical instruments, primarily supporting object localization tasks. Subsequent datasets, including SAR-RARP50~\cite{psychogyios2023sar}, PhaKIR~\cite{rueckert2025video}, and SurgVU~\cite{zia2025surgical}, introduce additional supervisory signals such as action labels and phase annotations to enrich surgical context. However, these contextual cues are not explicitly grounded to instance-level segmentation masks and therefore cannot directly support RA-SIS, which requires mapping query semantics to precise pixel-level targets. To support RA-SIS, a dedicated dataset should satisfy the following requirements: (i) large-scale surgical images covering realistic procedural variations; (ii) instance-level segmentation masks distinguishing multiple instruments and components; (iii) expert-driven, task-oriented query-answer pairs reflecting surgical reasoning; and (iv) explicit alignment between textual supervision and corresponding segmentation targets. Motivated by these principles, we construct the \textbf{Surgical Reasoning Segmentation Dataset (SurgRS)}, a benchmark specifically designed for task-driven reasoning-aware surgical instrument segmentation.

\begin{figure}[t]
\includegraphics[width=\textwidth]{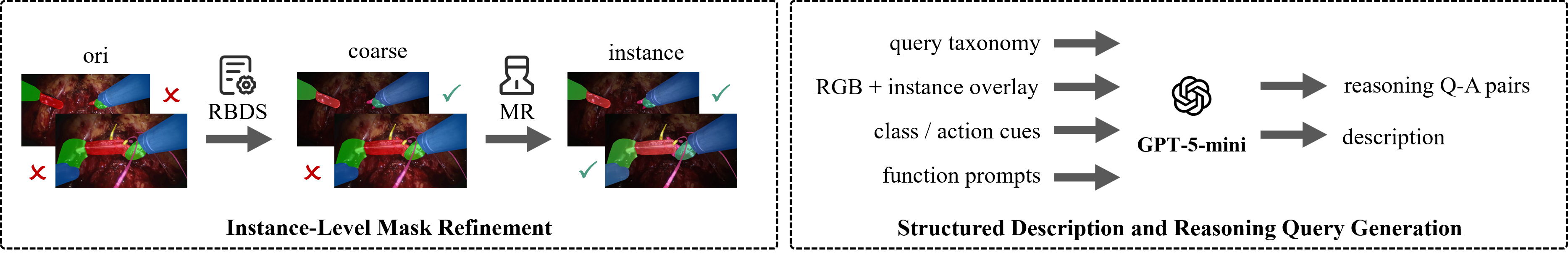}
\caption{Overall construction pipeline of SurgRS.} \label{fig:1}
\end{figure}

\subsubsection{SurgRS Construction Pipeline.}

SurgRS is constructed based on the SAR-RARP50 dataset~\cite{psychogyios2023sar}, which provides action labels offering task-driven procedural context conducive to reasoning-aware segmentation, as well as diverse and richly annotated surgical frames that support large-scale instance refinement. The overall construction pipeline of SurgRS is illustrated in Fig.~\ref{fig:1} and consists of two major stages.

\paragraph{(1) Instance-Level Mask Refinement.}

Following the preprocessing protocol of SAR-RARP50, we construct SurgRS from its RGB frames, semantic masks, and action annotations, and extract high-quality mask-annotated frames as the foundation of SurgRS. We remove 201 frames without valid targets and convert the remaining semantic masks into instance-level masks. Since semantic labels in SAR-RARP50 often merge multiple instrument instances, we refine them into instance-level masks. 

Specifically, we design rule-based decomposition strategies (RBDS) to split separable semantic masks into coarse instance masks, particularly for instruments with clear spatial separation and stable positional cues. RBDS performs connected-component analysis for each semantic tool region: large components are kept as instance candidates, while small regions caused by occlusion or thin structures are merged into the semantically closest component. For multi-instance tools, roles are assigned according to relative position and temporal continuity. Since RBDS only produces coarse masks, we then perform manual refinement (MR) in Supervisely to correct merged regions, resolve boundary ambiguities, and eliminate labeling errors. All refined masks are further verified to ensure structural correctness and annotation consistency.

\paragraph{(2) Structured Description and Reasoning Query Generation.}

As shown in Table~\ref{tab:1query_taxonomy}, we define nine representative reasoning query types arising in robotic-assisted prostate suturing, with an additional \emph{Other} (OT) category for rare but clinically valuable cases. This taxonomy constrains query design to clinically meaningful reasoning skills with strong intraoperative relevance.

To enhance clinical relevance and contextual accuracy, for each frame we use the vision-capable GPT-5-mini conditioned on the RGB image, labeled instance overlay, visible instrument categories, action annotation, and commonsense prompts about instrument functions. The model jointly generates a clinically natural frame description and 1-3 reasoning query-answer pairs according to the interaction complexity, consistent with prior reasoning segmentation frameworks~\cite{zhang2025pathmr,gong2025devil}. To enforce explicit reasoning-mask alignment, each answer must reference the corresponding segmentation targets using exact class names and attach a \texttt{<SEG>} marker. All query-answer pairs are manually checked with the RGB image and labeled overlay to ensure that the query is clinically reasonable and the target answer is correct.

\begin{table}[t]
\centering
\caption{SurgRS reasoning query taxonomy.}
\label{tab:1query_taxonomy}
\setlength{\tabcolsep}{3pt}
\renewcommand{\arraystretch}{1.05}
\resizebox{\linewidth}{!}{
\begin{tabular}{c|l|l}
\hline
\textbf{Code} & \textbf{Type} & \textbf{Definition} \\
\hline
LT  & Leading Tool & Identify the instrument executing the maneuver. \\
\hline
SE  & Support Exposure & Analyze exposure or counter-traction roles. \\
\hline
SGT & Suture Guide/Tension & Identify instruments guiding or tensioning sutures. \\
\hline
PAI  & Procedural Action Inference & Infer the current action from visible instruments. \\
\hline
NSI & Next-Step Importance & Identify instruments important for the next action. \\
\hline
NDL & Near/Deep Layout & Analyze spatial distribution in surgical field. \\
\hline
OR  & Occlusion Relation & Identify occlusion or overlap relationships. \\
\hline
PL  & Position Locate & Localize instruments using positional cues. \\
\hline
CR  & Commonsense Reference & Localize instruments via commonsense functional descriptions. \\
\hline
OT  & Other & Cover rare clinically meaningful cases. \\
\hline
\end{tabular}}
\end{table}

\subsection{SIRA Architecture}

\subsubsection{Overall Architecture.}

\begin{figure}[t]
\includegraphics[width=\textwidth]{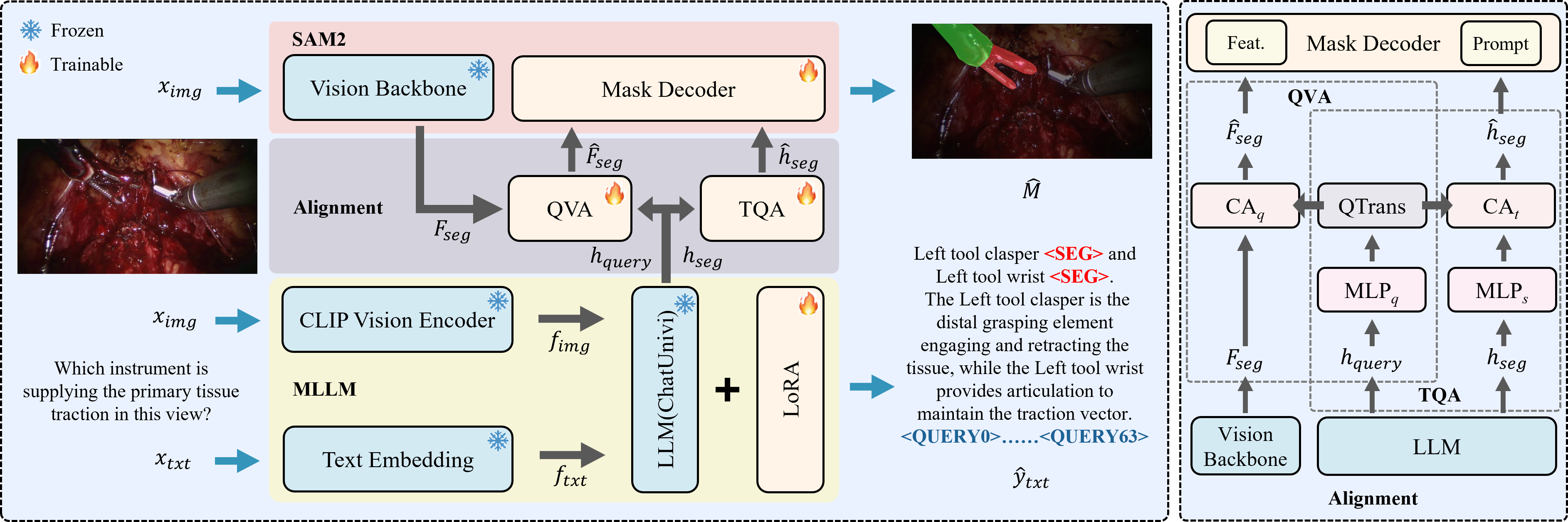}
\caption{Overview of the proposed SIRA framework.} \label{fig:2}
\end{figure}

As illustrated in Fig.~\ref{fig:2}, SIRA consists of a multimodal large language model (MLLM), a query-anchored dual alignment module, and a SAM2-based segmentation model~\cite{ravi2024sam2}.

The MLLM adopts Chat-UniVi~\cite{jin2024chat} with LoRA~\cite{hu2022lora} adapters for surgical domain adaptation. Given a textual query and an input image, it produces (i) target-level embeddings from \texttt{<SEG>} tokens, (ii) query-level embeddings from \texttt{<QUERY>} tokens, and (iii) explanatory text. The segmentation model follows SAM2, comprising a vision backbone and a mask decoder.

Since the MLLM and SAM2 operate in different representation spaces, and \texttt{<SEG>} prompts alone cannot encode full reasoning semantics, we introduce a query-anchored dual alignment module to bridge them. Query-level embeddings are projected into the visual feature space and injected into spatial features via cross-attention to produce query-conditioned representations. Target-level embeddings are projected into the SAM2 prompt space and further aligned with query semantics to enhance consistency between target activation and global reasoning context. The mask decoder then integrates aligned visual features and prompt embeddings to generate one or multiple masks under task-driven specifications. The entire framework is trained end-to-end with segmentation and text supervision.

\subsubsection{Disentangled Semantic Encoding.}

Existing reasoning-aware segmentation methods typically rely on a single dedicated \texttt{<SEG>} token to encode target-specific semantics from the MLLM, which entangles object activation and query-level reasoning within a unified representation. Such coupled encoding is insufficient for capturing the rich global context embedded in complex reasoning queries. To explicitly disentangle target-level activation from query-level semantics, we introduce a structured semantic encoding scheme that separates instance-oriented signals and coarse reasoning context in the latent space.

In addition to conventional \texttt{<SEG>} tokens indicating candidate segmentation targets, we incorporate a fixed-length set of $N_q$ learnable \texttt{<QUERY>} tokens to model global reasoning semantics independent of explicit instance activation.

Given a tokenized textual expression $x_{txt} = \{w_1, \dots, w_L\}$ and an image input $x_{img}$, the MLLM processes the multimodal input and autoregressively generates a response sequence $\hat{y}_{txt} = \{\hat{y}_1, \dots, \hat{y}_N\}$. The sequence $\hat{y}_{txt}$ includes explanatory tokens, multiple \texttt{<SEG>} tokens corresponding to potential targets, and a contiguous block of $N_q$ \texttt{<QUERY>} tokens representing query-level reasoning semantics. Let $\mathbf{H} \in \mathbb{R}^{N \times d}$ denote the hidden states of the final MLLM layer, where $d$ is the embedding dimension. We extract the representations associated with the \texttt{<SEG>} and \texttt{<QUERY>} tokens as

\begin{equation}
\boldsymbol{h}_{seg} \in \mathbb{R}^{K \times d}, \quad 
\boldsymbol{h}_{query} \in \mathbb{R}^{N_q \times d},
\end{equation}

\noindent where $K$ denotes the number of \texttt{<SEG>} tokens. Here, $\boldsymbol{h}_{seg}$ encodes instance-specific activation cues, while $\boldsymbol{h}_{query}$ captures disentangled coarse-grained reasoning context derived from $x_{txt}$ and $x_{img}$. This structured semantic decomposition establishes complementary instance-level and query-level representations for subsequent alignment.

\subsubsection{Query-Anchored Dual Alignment.}

Although $\boldsymbol{h}_{seg}$ provides instance-level activation cues, it does not fully preserve the coarse query semantics encoded in $\boldsymbol{h}_{query}$, which are essential for precise localization. As illustrated in Fig.~\ref{fig:2}, we introduce a query-anchored dual alignment mechanism to propagate disentangled query representations into spatial features and prompt embeddings.

\paragraph{(1) Query-to-Visual Alignment (QVA).}

Let $\boldsymbol{h}_{query} \in \mathbb{R}^{N_q \times d_m}$ denote the query-level embeddings extracted from the MLLM. Here, $N_q$ is the number of \texttt{<QUERY>} tokens and $d_m$ is the embedding dimension. Let $\boldsymbol{F}_{seg} \in \mathbb{R}^{HW \times d_v}$ denote the fine-grained spatial features extracted by the SAM2 backbone. We first project $\boldsymbol{h}_{query}$ into the vision feature space:

\begin{equation}
\boldsymbol{E}_{query}
=
\mathrm{QTrans}\big(\mathrm{MLP}_q(\boldsymbol{h}_{query})\big),
\end{equation}

\noindent where $\boldsymbol{E}_{query} \in \mathbb{R}^{N_q \times d_v}$ encodes query semantics in the visual feature dimension, and $\mathrm{QTrans}(\cdot)$ denotes a lightweight transformer block for modeling inter-query dependencies. We then align query semantics with spatial representations via cross-attention:

\begin{equation}
\hat{\boldsymbol{F}}_{seg}
=
\mathrm{CrossAttention}_{q}
(
\boldsymbol{F}_{seg}^{Q},
\boldsymbol{E}_{query}^{K},
\boldsymbol{E}_{query}^{V}
),
\end{equation}

\noindent where spatial features act as queries, and query embeddings serve as keys and values. The resulting $\hat{\boldsymbol{F}}_{seg} \in \mathbb{R}^{HW \times d_v}$ represents query-conditioned visual features.

\paragraph{(2) Target-to-Query Alignment (TQA).}

The instance-level embeddings $\boldsymbol{h}_{seg} \in \mathbb{R}^{K \times d_m}$ are first projected into the segmentation prompt embedding space:

\begin{equation}
\tilde{\boldsymbol{h}}_{seg}
=
\mathrm{MLP}_s(\boldsymbol{h}_{seg}),
\end{equation}

\noindent where $\tilde{\boldsymbol{h}}_{seg} \in \mathbb{R}^{K \times d_p}$ and $d_p$ denotes the SAM2 prompt embedding dimension. To anchor target activation within shared query semantics, we further align the projected target embeddings with the query representations:

\begin{equation}
\hat{\boldsymbol{h}}_{seg}
=
\mathrm{CrossAttention}_{t}
(
\tilde{\boldsymbol{h}}_{seg}^{Q},
\boldsymbol{E}_{query}^{K},
\boldsymbol{E}_{query}^{V}
).
\end{equation}

Through QVA and TQA, both spatial features and target prompts are conditioned on the same disentangled query representation, establishing a shared semantic anchor across modalities.

\subsubsection{Training Objective.}

The aligned visual features $\hat{\boldsymbol{F}}_{seg}$ and target prompt embeddings $\hat{\boldsymbol{h}}_{seg}$ are fed into the SAM2 mask decoder to produce one binary mask per target, naturally supporting multi-target prediction under a single query.

SIRA is trained end-to-end with both segmentation and text generation supervision. The overall objective is defined as

\begin{equation}
\mathcal{L}_{total}
=
\lambda_{bce}\mathcal{L}_{bce}
+
\lambda_{dice}\mathcal{L}_{dice}
+
\lambda_{txt}\mathcal{L}_{txt},
\end{equation}

\noindent where $\mathcal{L}_{bce}$ and $\mathcal{L}_{dice}$ denote the binary cross-entropy and Dice losses for mask prediction, respectively, and $\mathcal{L}_{txt}$ is the cross-entropy loss for autoregressive text generation.

\section{Experiments}

\subsubsection{Dataset.}

We evaluate SIRA on the proposed SurgRS dataset. SurgRS is constructed based on SAR-RARP50~\cite{psychogyios2023sar} and focuses exclusively on radical prostatectomy procedures. The dataset contains approximately 16,000 surgical images (resolution $1920 \times 1080$) and around 41,000 structured image-text pairs, covering 20 instrument categories and over 127,000 instance-level segmentation targets. Following the original SAR-RARP50 protocol, we split the dataset into training and testing sets with a ratio of 4:1.

\subsubsection{Implementation Details.}

We fine-tune the MLLM (Chat-UniVi-7B) using LoRA with rank 8 while freezing the remaining language model parameters. The number of query tokens is set to $N_q = 64$. The alignment module (ViT, cross-attention, and MLP projections) and the SAM2 mask decoder are trained, while freezing the SAM2 vision backbone and other components. Training is performed using the AdamW optimizer with a learning rate of 3e-4 and no weight decay. We adopt a WarmupDecayLR scheduler with 100 warm-up iterations. The model is trained for 10 epochs (13,000 iterations per epoch) on the SurgRS training set. The segmentation loss weights are set to $\lambda_{bce}=2$ and $\lambda_{dice}=0.5$, and the reasoning loss weight is $\lambda_{txt}=1$. Experiments are conducted on two NVIDIA GeForce RTX 3090 GPUs using DeepSpeed. The per-GPU batch size is 1 with gradient accumulation set to 1, resulting in a total batch size of 2. The training process consumes approximately 24GB of GPU memory, while inference requires about 16GB.

\subsubsection{Baselines and Evaluation Metrics.}

We compare SIRA with four state-of-the-art reasoning segmentation models, LISA~\cite{lai2024lisa}, GSVA~\cite{xia2024gsva}, PathMR~\cite{zhang2025pathmr}, and VRS-HQ~\cite{gong2025devil}, all implemented with 7B-scale language backbones for fair comparison. Following prior work, we adopt generalized IoU (gIoU) and complete IoU (cIoU) for evaluation. In addition, we report the Dice coefficient, which is widely used in medical image segmentation.

\begin{table*}[t]
\centering
\caption{Quantitative comparison of reasoning segmentation performance on SurgRS (\% for gIoU, cIoU, and Dice).}
\label{tab:2results}
\setlength{\tabcolsep}{4pt}
\renewcommand{\arraystretch}{1.15}
\resizebox{\textwidth}{!}{
\begin{tabular}{l|ccc|cccccccccc}
\hline
\multirow{2}{*}{Method} & \multicolumn{3}{c|}{Overall} & \multicolumn{10}{c}{Reasoning Query Type} \\
\cline{2-4}\cline{5-14}
 & gIoU & cIoU & Dice & LT & SE & SGT & PAI & NSI & NDL & OR & PL & CR & OT \\
\hline
LISA     & 65.05 & 74.32 & 74.21 & 79.34 & 71.14 & 69.24 & 73.23 & 74.55 & 74.43 & 68.30 & 70.86 & 58.81 & 54.35 \\
GSVA     & 65.30 & 75.08 & 74.39 & 79.14 & 72.20 & 69.52 & 73.03 & 74.52 & 74.89 & 67.21 & 69.64 & 60.68 & 53.94 \\
PathMR   & 66.42 & 75.14 & 76.25 & 80.18 & 74.11 & 72.39 & 75.91 & 76.40 & 75.69 & 71.15 & 73.37 & 63.20 & 61.10 \\
VRS-HQ   & 70.79 & 80.60 & 79.61 & 83.53 & 78.57 & 75.40 & 78.53 & 78.92 & 79.78 & 72.31 & 76.29 & 64.97 & 59.82 \\
\hline
\textbf{SIRA} & \textbf{72.24} & \textbf{81.78} & \textbf{80.93}
& \textbf{86.82} & \textbf{80.05} & \textbf{76.76} & \textbf{79.53} & \textbf{80.02} & \textbf{81.05} & \textbf{74.42} & \textbf{76.78} & \textbf{65.58} & \textbf{61.61} \\
\hline
\end{tabular}}
\end{table*}

\begin{figure}[t]
\includegraphics[width=\textwidth]{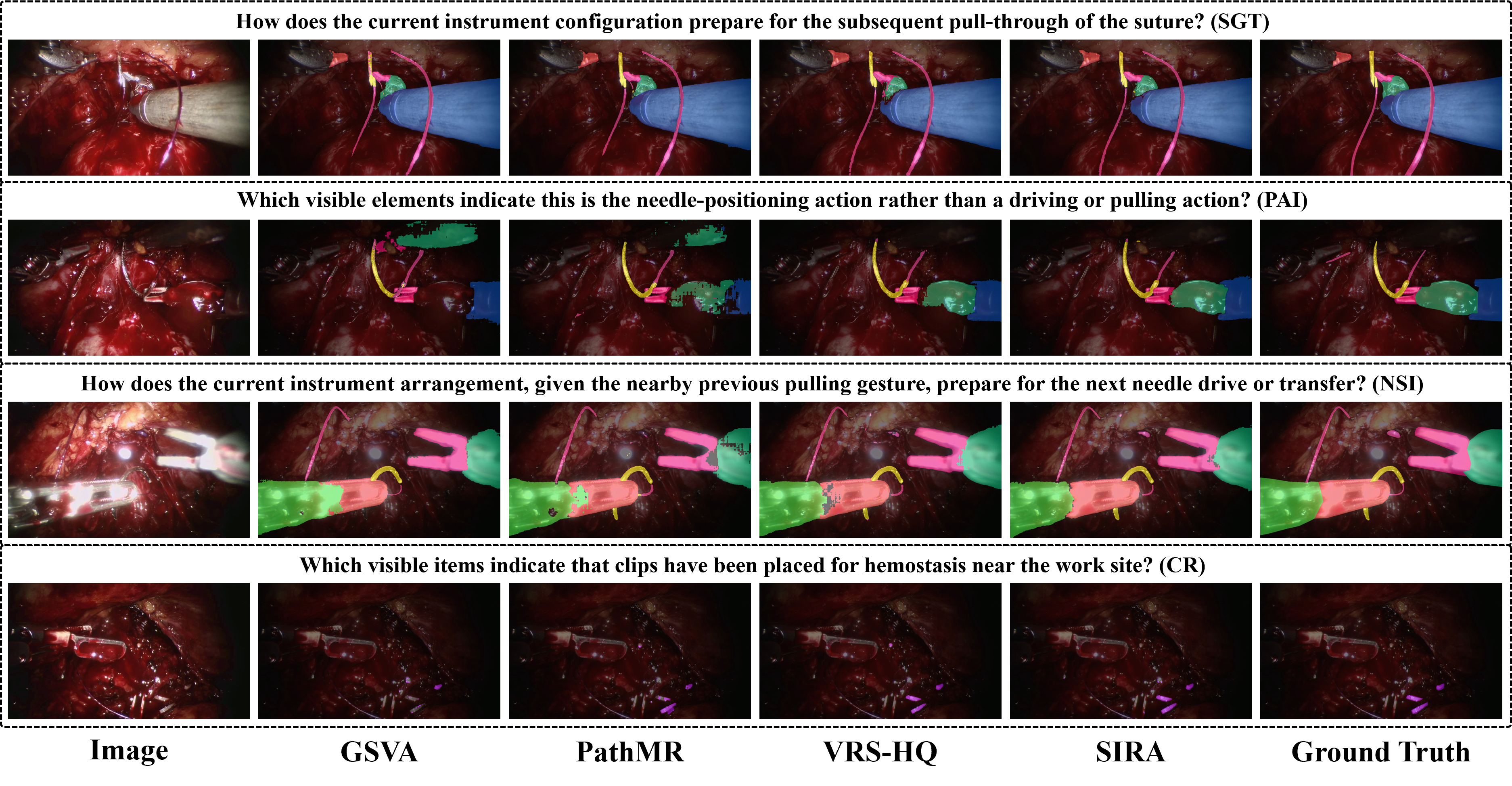}
\caption{Exemplar qualitative results of different approaches on SurgRS.} \label{fig:3vis}
\end{figure}

\subsubsection{Comparison Results.}

Table~\ref{tab:2results} reports the reasoning segmentation performance on SurgRS. SIRA achieves the best results across all overall metrics and reasoning query types. Fig.~\ref{fig:3vis} illustrates qualitative comparisons under several challenging surgical conditions. In scenarios involving ambiguous boundaries and small-scale instruments, SIRA produces more complete and spatially coherent masks, with clearer delineation of fine structures. Under low-illumination environments and for infrequently occurring targets, the proposed method maintains stable localization, whereas competing methods exhibit partial activation or omission. Furthermore, in cases affected by motion blur due to rapid instrument movement, SIRA preserves consistent mask prediction and reduces fragmentation artifacts. These observations indicate that the proposed query-anchored dual alignment mechanism enhances semantic-visual consistency and improves robustness across diverse intraoperative challenges.

\subsubsection{Ablation Analysis.}

Table~\ref{tab:3ablation} reports ablations of QVA and TQA. Enabling either module consistently improves over the baseline, with QVA providing slightly larger gains. Combining QVA and TQA yields the best results, indicating that the two alignments are complementary for query-conditioned surgical instrument segmentation.

\begin{table}[t]
\centering
\caption{Ablation study of QVA and TQA on SurgRS (\%).}
\label{tab:3ablation}
\setlength{\tabcolsep}{6pt}
\renewcommand{\arraystretch}{1.2}
\begin{tabular}{c c c | c c c}
\hline
Baseline & QVA & TQA & gIoU & cIoU & Dice \\
\hline
$\bullet$ & $\circ$ & $\circ$ & 70.79 & 80.60 & 79.61 \\
$\bullet$ & $\circ$ & $\bullet$ & 71.40 & 81.23 & 80.09 \\
$\bullet$ & $\bullet$ & $\circ$ & 71.93 & 81.62 & 80.59 \\
\hline
$\bullet$ & $\bullet$ & $\bullet$ & \textbf{72.24} & \textbf{81.78} & \textbf{80.93} \\
\hline
\end{tabular}
\end{table}

\section{Conclusion}

We introduce RA-SIS, a reasoning-aware surgical instrument segmentation formulation that models segmentation as query-conditioned inference under procedural context. To benchmark this setting, we construct SurgRS, a dataset aligning instance-level masks with structured query-answer supervision for pixel-level semantic grounding. We further propose SIRA, a multimodal framework that disentangles query- and target-level semantics and aligns them with visual features via query-anchored dual alignment. Experiments demonstrate consistent improvements over state-of-the-art reasoning segmentation methods on SurgRS. Future work will extend RA-SIS to video-based surgical understanding, where temporal consistency and motion dynamics introduce additional challenges.

\subsubsection{Disclosure of Interests.}

The authors have no competing interests to declare that are relevant to the content of this article.

\bibliographystyle{unsrt}
\bibliography{Paper-1772}

\end{document}